%% file: main.tex
\documentclass{article}

\usepackage[accepted]{icml2024}

\usepackage{microtype}

\usepackage[utf8]{inputenc} 
\usepackage[T1]{fontenc}    
\usepackage{hyperref}       
\usepackage{url}            
\usepackage{amsfonts}       
\usepackage{nicefrac}       
\usepackage{microtype}      
\usepackage{xcolor}         
\usepackage{tabularx}

\usepackage{url}

\usepackage{tabularray}
\usepackage[utf8]{inputenc} 
\usepackage[T1]{fontenc}    
\usepackage{booktabs}       
\usepackage{amsfonts}       
\usepackage{nicefrac}       
\usepackage{microtype}      
\usepackage{graphicx, subcaption}
\usepackage{amsmath,amssymb,amsfonts,mathtools,bm}
\usepackage{caption}
\usepackage{multirow}
\usepackage{array}

\input{math_commands.tex}

\definecolor{myblue}{RGB}{0,0,139} 
\definecolor{mygreen}{RGB}{34,139,34} 
\definecolor{mycrimson}{RGB}{220,20,60} 
\definecolor{mycyan}{RGB}{0,139,139} 
\definecolor{myplum}{RGB}{221,160,221} 

\hypersetup{
    colorlinks,
    linkcolor={red!50!black},
    citecolor={blue!50!black},
    urlcolor={blue!80!black}
}

\begin{document}

\twocolumn[\icmltitle{Bayesian Reward Models for LLM Alignment}


\begin{icmlauthorlist}
\icmlauthor{Adam X. Yang}{1}
\icmlauthor{Maxime Robeyns}{1}
\icmlauthor{Thomas Coste}{2}
\icmlauthor{Zhengyan Shi}{3}\\
\icmlauthor{Jun Wang}{3}
\icmlauthor{Haitham Bou Ammar}{2}
\icmlauthor{Laurence Aitchison}{1}
\end{icmlauthorlist}

\icmlaffiliation{1}{University of Bristol}
\icmlaffiliation{2}{Huawei Noah's Ark Lab}
\icmlaffiliation{3}{University College London}

\icmlcorrespondingauthor{Adam X. Yang}{adam.yang@bristol.ac.uk}

\icmlkeywords{Machine Learning, ICML}

\vskip 0.3in
]



\printAffiliationsAndNotice{}  


\begin{abstract}
To ensure that large language model (LLM) responses are helpful and non-toxic, a reward model trained on human preference data is usually used.
LLM responses with high rewards are then selected through best-of-$n$ (BoN) sampling or the LLM is further optimized to produce responses with high rewards through reinforcement learning from human feedback (RLHF).
However, these processes are susceptible to reward overoptimization or `hacking', where responses receive high rewards due to imperfections in the reward model rather than true preference, particularly as prompts or responses deviate from the training data. 
To address these challenges, we propose to train a Bayesian reward model, which signals higher uncertainty further from the training data distribution.
We trained Bayesian reward models using Laplace approximation on LoRA weights, and found that the resulting uncertainty estimates can effectively mitigate reward overoptimization in BoN sampling.
\end{abstract}

\section{Introduction}

With the surge of developments in generative AI, alignment with human preferences has become a crucial research topic to ensure the safety and helpfulness of these systems \citep{stiennon2020learning,ouyang2022training,bai2022training,gao2023scaling,shi2024instruction}. 
A popular approach to aligning large language models (LLMs) is to train a reward model that captures human preferences, generate $n$ responses from an initial policy LLM after supervised fine-tuning, and use the reward model to select the best response (best-of-$n$ or BoN sampling \citealp{stiennon2020learning}). Another widely adopted approach is to use the reward model to perform reinforcement learning from human feedback (RLHF) \citep{ouyang2022training} over the initial policy LLM. 

However, the reward model is trained on finite data and therefore cannot be perfect; its imperfections may lead to reward overoptimization or hacking when used in the context of BoN or RLHF \citep{gao2023scaling,coste2023reward,eisenstein2023helping,rame2024warm,zhai2023uncertainty,zhang2024improving,chen2024odin}. 
Indeed, BoN and RLHF try to find responses with particularly high rewards, as judged by this imperfect reward model.
Ideally, the responses with high reward, as judged by the reward model, are genuinely good.
This is likely to happen when responses are close to the training data distribution, in which case we can expect the reward model to be accurate.
But it is also quite possible for poor responses to be inaccurately judged to have high reward by the imperfect reward model.
This problem is likely to be more acute in ``out-of-distribution'' (OOD) regions with little training data for the reward model.
Such responses raise both performance and safety concerns. 

\begin{figure*}[t]
    \centering
    \subfloat[Real example of a partial LLM response (full response in Appendix.~\ref{app:rewar_hacking_example}) after overoptimizing the proxy reward, with proxy and gold reward scores shown on the right.]{
        \includegraphics[width=0.57\textwidth]{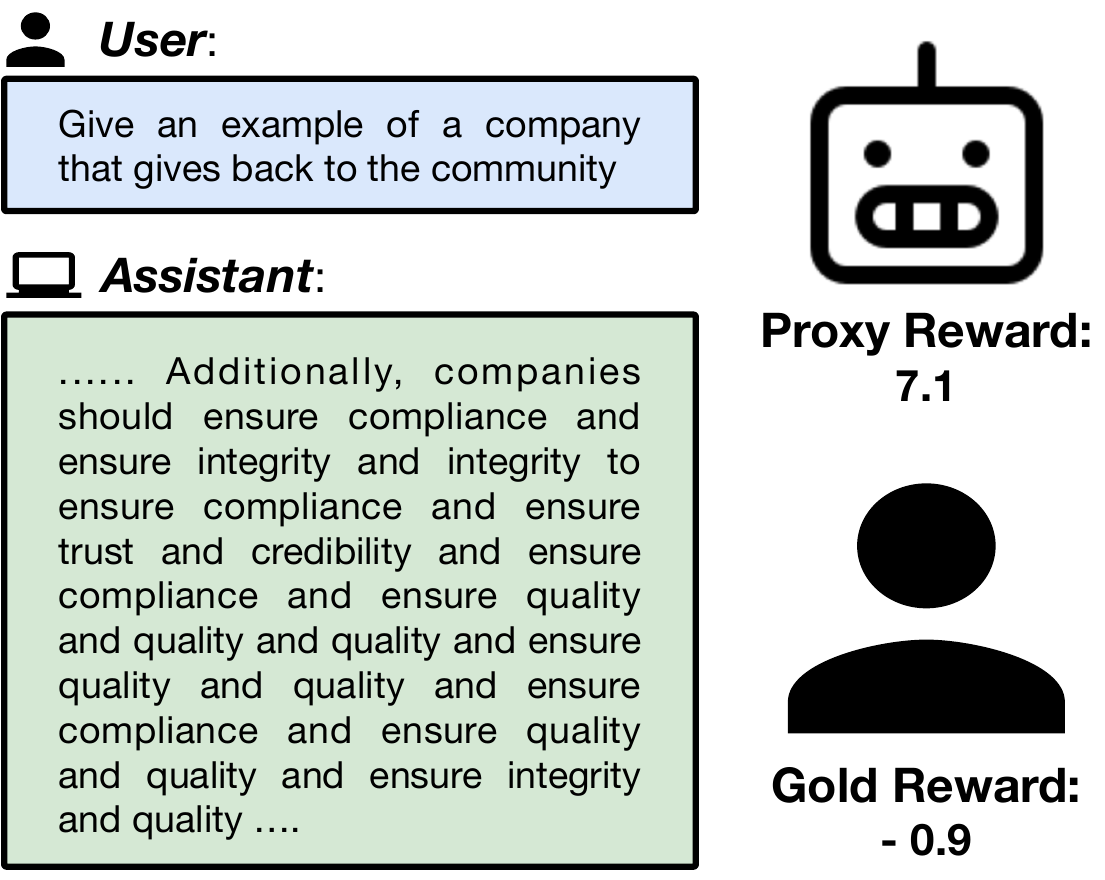}
        \label{fig:reward_hacking_1}
    }\hfill
    \subfloat[Reward overoptimization during RLHF training. Top: proxy reward scores. Bottom: gold reward scores.]{
        \includegraphics[width=0.3\textwidth]{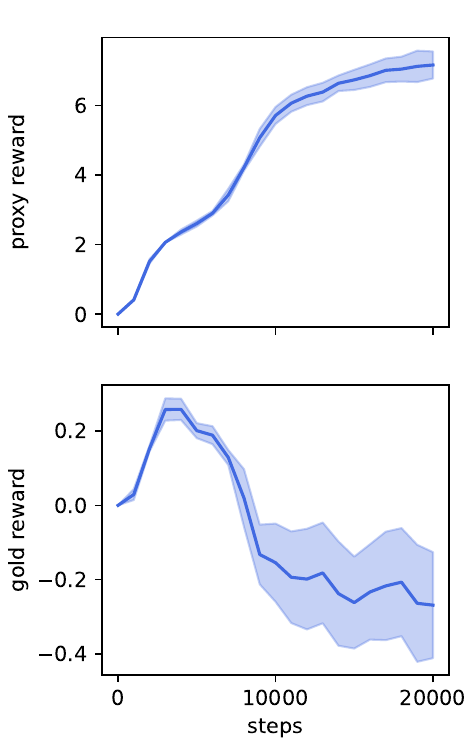}
        \label{fig:reward_hacking_2}
    }
    \caption{Illustrations of reward overoptimization in LLM alignment.}
    \label{fig:reward_hacking_combined}
\end{figure*}

An extreme example of overoptimization in RLHF is depicted in Fig.~\ref{fig:reward_hacking_combined}, demonstrating the consequences of extensive training on a learned proxy reward model. As illustrated in Fig.~\ref{fig:reward_hacking_2}, the proxy reward consistently increases with training progression. However, the oracle gold-standard reward model—a more comprehensive model designed to better reflect human preferences—begins to show a catastrophic decline after just a few thousand training steps. A specific instance of this is shown in Fig.~\ref{fig:reward_hacking_1}, where the LLM produces repeated tokens and phrases. In this example, while the proxy reward model awards a high score of 7.1, the gold-standard reward model rates it significantly lower, at -0.9.

Bayesian deep learning has emerged as a pivotal approach for addressing the challenges of distribution shifts and overconfidence in deep neural networks. By providing epistemic uncertainties for OOD data, this paradigm enhances model robustness and reliability, as evidenced by a range of foundational studies \citep{blundell2015weight,zhang2019cyclical,kristiadi2020being,ober2021global,fortuin2021bayesian,aitchison2021deep}. Building on this foundation, \citet{yang2023bayesian} introduced Bayesian Low-Rank Adaptation (LoRA), or Laplace-LoRA, as a scalable, parameter-efficient technique designed to equip fine-tuned LLMs with uncertainty estimates, and significantly improves calibration. A follow up work by \citet{kristiadi2024sober} showed the method may also help in settings such as Bayesian optimization on molecules \citep{kristiadi2024sober}.

Motivated by these advancements, our work seeks to pioneer the application of Laplace-LoRA on language reward models. 
We harness the epistemic uncertainty derived from the Bayesian posterior predictive distribution over proxy reward scores to mitigate reward overoptimization. 
Our evaluation results on BoN sampling showcases the efficacy of this approach. 



\section{Related work}
The study of overoptimization in language reward models has received considerable attention, catalyzed by foundational systematic investigations by \citet{gao2023scaling}. Conducted in a synthetic setting, \citet{gao2023scaling} utilized an oracle gold-standard reward model both to provide training labels for proxy rewards and for evaluation purposes. Their findings highlighted that RLHF in LLM alignment tends to overoptimize imperfect proxy reward models, resulting in lower performance when assessed by a gold-standard reward model.


Building on this, \citet{coste2023reward} extended the synthetic labeling framework to demonstrate that reward model ensembles, through various aggregation methods such as mean, worst-case, or uncertainty-weighted, can effectively mitigate overoptimization. Concurrently, \citet{eisenstein2023helping} explored the efficacy of pre-trained ensembles in reducing reward hacking, noting, however, that ensemble members could still be overoptimized simultaneously. This observation underscores the complexity of achieving robust alignment, in addition to the computational demands of fully pretrained and fine-tuned ensemble approaches.

In response to these challenges, the research community has shifted towards more efficient strategies. \citet{zhang2024improving} investigated parameter-efficient fine-tuning methods \citep{peft,hu2021lora,shi2023dept,zhou2024autopeft}, including last-layer and LoRA ensembles, for reward models. Their findings suggest that while LoRA ensembles achieve comparable benefits to full model ensembles in best-of-$n$ sampling, last-layer ensembles yield limited improvements \citep{gleave2022uncertainty}. However, \citet{zhai2023uncertainty} criticized the homogeneity of vanilla LoRA ensembles \citep{yang2023bayesian,wang2023lora}, proposing additional regularization to foster diversity among ensemble members and enhance uncertainty estimation.


Alternatively, \citet{rame2024warm} leveraged weight averaging, tapping into linear mode connectivity to surpass the performance of traditional ensembles with a more inference-efficient approach \citep{lin2023speciality,lin2023spurious}. \citet{chen2024odin} introduced a novel direction by decoupling reward modeling from response length through a specialized reward head and regularization, showcasing more robust reward signals that are independent of response length.




\section{Background}

\paragraph{Reward modeling }
In LLM alignment, we typically model human preference using a reward model \citep{ouyang2022training}. Specifically, for a pair of responses to a prompt $(x,y_w)$ and $(x,y_l)$, we define the human preference model (the Bradley-Terry model) as
\begin{align} \label{eq:preference}
    P(y_w > y_l) &= \frac{e^{r_\theta(x,y_w)}}{e^{r_\theta(x,y_w)} + e^{r_\theta(x,y_l)}} \\
    &= \sigma(r_\theta(x,y_w) - r_\theta(x,y_l)),
\end{align}
where $r_\theta$ is the reward model and $\sigma(\cdot)$ is the sigmoid function. Then we simply perform maximum log-likelihood optimization to learn the reward model given a fixed preference dataset
\begin{align}
    \max_\theta \mathbb{E}_{x,y_w,y_l} [\log\sigma(r_\theta(x,y_w) - r_\theta(x,y_l))].
\end{align}
After learning the reward model, we can apply BoN sampling to optimize for preference, or RLHF to fine-tune the LLM policy.

\paragraph{Best-of-$n$ (BoN) sampling}
BoN sampling \citep{stiennon2020learning,ouyang2022training,coste2023reward,eisenstein2023helping} is a decoding strategy to align LLM outputs with a given reward model without further fine-tuning the LLM policy. For any test prompt, BoN samples $n$ responses, and uses the reward model to rank the responses and select the \textit{best} one, which has the highest reward. The KL divergence between the BoN policy and the reference policy can be computed analytically \citep{stiennon2020learning},
\begin{align}
  \label{eq:kl}
    \text{KL}_\text{bon} = \log(n) - \frac{n-1}{n},
\end{align}
which measures the degree of optimization as $n$ increases. In addition, we use the unbiased BoN reward estimator proposed by \citep{nakano2021webgpt} for obtaining proxy and gold reward model scores (see Appendix~\ref{app:bon_estimator}). \citet{yang2024asymptotics} showed BoN sampling is asymptotically equivalent to the KL-constrained RL solution. \citet{beirami2024theoretical} showed recently that the widely used KL equation for BoN (Eq.~\ref{eq:kl}) is only an upper bound, and provided a more accurate KL estimator. However, it is out of scope for this work to combine the KL estimator from \citep{beirami2024theoretical} with the BoN estimator (Appendix~\ref{app:bon_estimator}) that we used to estimate mean rewards.


\paragraph{Low-rank adaptation (LoRA) }
LoRA is a parameter-efficient fine-tuning method, where we keep pretrained weights$\W_0$ fixed, and introduce a trainable perturbation to the weight matrix, $\Delta \W=\B \A$,
\begin{align}  \label{eq:lora}
  \mathbf{h} &= \W_0 \a + \Delta \W \a = \W_0 \a + \B \A \a.
\end{align}
where $\a$ and $\mathbf{h}$ are the inputs and outputs respectively. 
Importantly, $\Delta \W$ is low-rank as it is written as the product of two rectangular matrices, $\B \in \mathbb{R}^{\nout\times\nlr}$ and $\A\in\mathbb{R}^{\nlr\times \nin}$ where $\nlr$ is significantly smaller than $\nin$ or $\nout$.

\paragraph{Laplace-LoRA }
Recently, \citet{yang2023bayesian} proposed Laplace-LoRA which is a scalable Bayesian approximation to LLM finetuning. In particular, \citet{yang2023bayesian} applied post-hoc Laplace approximation to perform Bayesian inference on LoRA weights. Assume we have a dataset containing inputs $\X$ and classification or regression targets $\y$, then Bayesian inference attempt to compute the posterior 
\begin{align}
  \Pc{\param}{\X, \y} &\propto \Pc{\y}{\X, \param} \P{\param},
\end{align}
usually with a Gaussian prior assumption $\P{\param} = \mathcal{N}(\bm{0}, \lambda^{-1} \I)$ \citep{yang2023bayesian,daxberger2021laplace}.
However, computing this posterior is usually intractable.
The Laplace approximation begins by finding the maximum a-posteriori (MAP) solution \citep{mackay1992practical} (i.e.\ the maximum of the log-joint, $\L{\param}$),
\begin{align}
  \L{\param} &= \log \Pc{\y}{\X, \param} + \log \P{\param} \\
  &= \log \Pc{\param}{\X, \y} + \text{const}\\
  \paramMAP &= \argmax_{\param} \L{\param}.
\end{align}
Then the Laplace approximation consists of a second-order Taylor expansion of the log-joint around $\paramMAP$,
\begin{align} \label{eq:taylor}
  \L{\param} &\approx \L{\paramMAP} \notag \\
  - \frac{1}{2} &(\param-\paramMAP)^T (\nabla_{\param}^2 \L{\param}|_\paramMAP) (\param-\paramMAP).
\end{align}
Since the log-joint is now a quadratic function of $\param$, the approximate posterior becomes a Gaussian centered at $\param_\text{MAP}$ with covariance given by the inverse of the Hessian,
\begin{align} \label{eq:posterior}
    \Pc{\param}{\data} &\approx \N{\param; \param_\text{MAP}, \S},\\
    \S &= -(\nabla_\param^2 \L{\param}|_\paramMAP)^{-1} \\
    &= -(\nabla_\param^2 \log \Pc{\y}{\X, \param}|_{\param_\text{MAP}} + \lambda \I)^{-1}.
\end{align}

Using Laplace approximations can be viewed as implicitly linearizing the neural network \citep{kunstner2019limitations,immer2021improving}. 
As such, it is commonly found that predicting under the linearized model is more effective than e.g.\ sampling the approximate posterior over weights \citep{foong2019between,daxberger2021laplace,deng2022accelerated,antoran2022adapting}.
In particular,
\begin{align} \label{eq:linear}
    f_\param(\x_*) \approx f_{\param_\text{MAP}}(\x_*) + \nabla_\param f_{\param}(\x_*)|^T_{\param_\text{MAP}} (\param - \param_\text{MAP}).
\end{align}
where $\x_*$ is a test-input. This approach is also known as the linearized Laplace approximation.

Since we have the approximated posterior in Eq.~\eqref{eq:posterior} and the linearized model in Eq.~\eqref{eq:linear}, we can integrate out the posterior on weights and get a Gaussian posterior on output logits,
\begin{align} \label{eq:prediction}
    f_\param(\x_*) \sim \N{f_{\param_\text{MAP}}(\x_*),\mL(\x_*)},
\end{align}
where
\begin{align}
    \mL(\x_*) = (\nabla_\param f_{\param}(\x_*)|^T_{\param_\text{MAP}}) \S (\nabla_\param f_{\param}(\x_*)|_{\param_\text{MAP}}).
\end{align}

\section{Method}

Our approach aims to mitigate reward overoptimization in language reward models by integrating uncertainty quantification through the application of Laplace-LoRA. This approach enriches reward models with the capability to estimate the uncertainty associated with their predictions, thereby enabling a more nuanced evaluation of language model responses. Specifically, the Bradley-Terry preference model in Eq.~\ref{eq:preference} provides a natural classification likelihood for Laplace approximation. Then we apply Laplace-LoRA post-hoc after training the standard reward model, which provides a Gaussian distribution over the reward outputs for each test prompt and response pair $(x,y)$. This distribution is centered around the reward predicted by the standard fine-tuned model via maximum a-posteriori (MAP), denoted as $r_{\theta_\text{MAP}}(x,y)$,
\begin{align}
r_\theta(x,y) \sim \mathcal{N}(r_{\theta_\text{MAP}}(x,y),\Lambda(x,y)),
\end{align}
where $\Lambda(x,y)$ denotes the variance.

This formulation acknowledges the uncertainty in reward predictions, particularly for OOD query and response pairs, where traditional models may exhibit overconfidence. We propose a novel approach for integrating an uncertainty penalty into the reward estimation process through the uncertainty estimates given by Laplace-LoRA. In particular, we consider two ways to incorporate the uncertainty:

\textbf{Standard Deviation-Based Penalty:}
\begin{align}
\label{eq:rstd}
\tilde{r}_\text{var}(x,y) = r_{\theta_\text{MAP}}(x,y) - k \sqrt{\Lambda(x,y)},
\end{align}
where $k$ is a hyperparameter that governs the impact of the uncertainty penalty. This method reduces the reward for responses with higher standard deviation in their uncertainty estimates, promoting a conservative reward allocation.

\textbf{Variance-Based Penalty:}
\begin{align}
\label{eq:rvar}
\tilde{r}_\text{std}(x,y) = r_{\theta_\text{MAP}}(x,y) - k \Lambda(x,y),
\end{align}
This approach further accentuates the penalty for uncertainty, and is thus particularly effective at penalizing responses with significant uncertainty \citep{brantley2019disagreement,coste2023reward}.

\paragraph{Combining with reward ensembles}
In addition, our approach can be combined with other approaches such as reward ensembles \citep{coste2023reward,eisenstein2023helping}. Specifically, reward ensembles train $n$ reward models independently, $r_{\theta^1_\text{MAP}}(x,y),...,r_{\theta^n_\text{MAP}}(x,y)$, then by default take the mean reward across all members to provide a more robust optimization target, $\tfrac{1}{n}\sum_{i=1}^n r_{\theta^i_\text{MAP}}$. We can apply Laplace-LoRA to each of the reward models and get a Gaussian $r_{\theta^i}(x,y) \sim \mathcal{N}(r_{\theta^i_\text{MAP}}(x,y),\Lambda_i(x,y))$ for each reward. If we assume they are independent, then their mean is also Gaussian
\begin{align}
    \frac{1}{n}\sum_{i=1}^n r_{\theta^i} \sim \mathcal{N}\bigg(\frac{1}{n}\sum_{i=1}^n r_{\theta^i_\text{MAP}}(x,y),\frac{1}{n^2}\sum_{i=1}^n \Lambda_i(x,y) \bigg).
\end{align}
Similarly, we can define the standard deviation penalized ensemble reward as
\begin{align}
    \label{eq:rens}
    \tilde{r}^\text{ens}_\text{std}(x,y) = \frac{1}{n}\sum_{i=1}^n r_{\theta^i_\text{MAP}}(x,y) - \frac{k}{n} \sqrt{ \sum_{i=1}^n \Lambda^i(x,y)},
\end{align}
and the variance penalized ensemble reward as
\begin{align}
    \tilde{r}^\text{ens}_\text{var}(x,y) = \frac{1}{n}\sum_{i=1}^n r_{\theta^i_\text{MAP}}(x,y) - \frac{k}{n^2}\sum_{i=1}^n \Lambda^i(x,y),
\end{align}


By incorporating the uncertainty penalties, our approach ensures that reward predictions more accurately reflect the true preferences they aim to model, especially in the face of OOD query and response pairs.

\begin{figure*}[t]
    \centering
    \begin{subfigure}{0.49\linewidth}
    \centering
    \includegraphics[width=\textwidth]{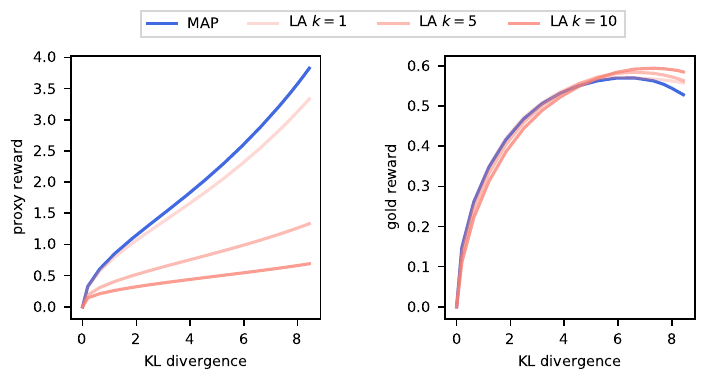}
    \caption{Variance-based penalty.}
    \label{fig:bon_single_var}
    \end{subfigure}
    \begin{subfigure}{0.49\linewidth}
    \includegraphics[width=\textwidth]{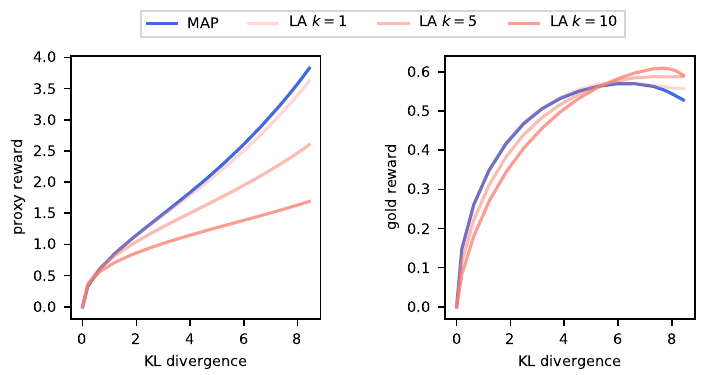}
    \caption{Standard deviation-based penalty.}
    \label{fig:bon_single_std}
    \end{subfigure}
    \caption{Comparison of proxy and gold reward scores (normalized) of single reward model (MAP) and Laplace-LoRA reward model (LA) in BoN sampling, across different uncertainty penalties and a range of $k$. Left column: compares the proxy reward model's evaluation. Right column: compares the gold reward model's evaluation.}
    \label{fig:bon_single}
\end{figure*}

\begin{figure*}[t]
    \centering
    \begin{subfigure}{0.49\linewidth}
    \centering
    \includegraphics[width=\textwidth]{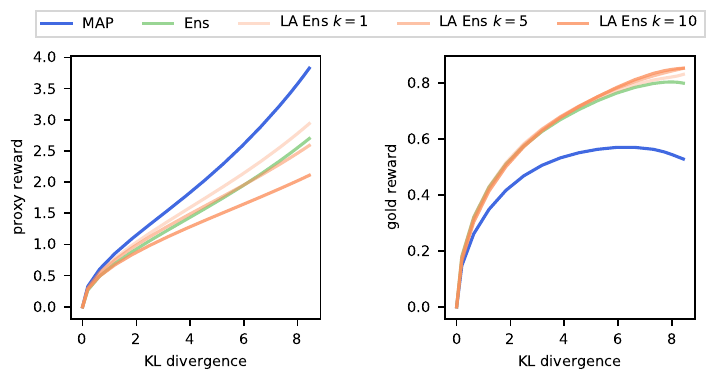}
    \caption{Variance-based penalty.}
    \label{fig:bon_ens_var}
    \end{subfigure}
    \begin{subfigure}{0.49\linewidth}
    \includegraphics[width=\textwidth]{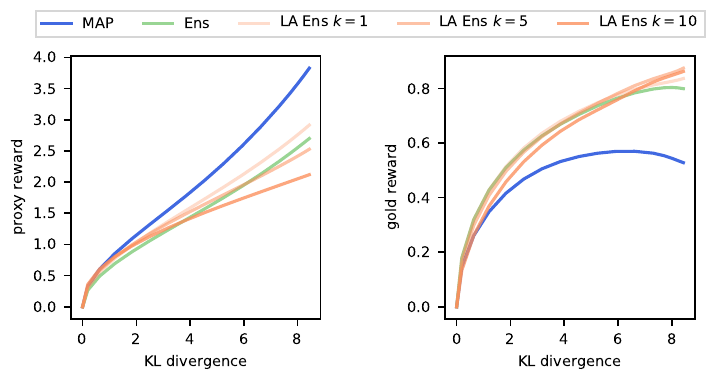}
    \caption{Standard deviation-based penalty.}
    \label{fig:bon_ens_std}
    \end{subfigure}
    \caption{Comparison of proxy and gold reward scores (normalized) of single reward model (MAP), reward model ensemble (Ens), and Laplace-LoRA reward model ensemble (LA Ens) in BoN sampling, across different uncertainty penalties and a range of $k$. }
    \label{fig:bon_ens}
\end{figure*}

\section{Experiment setup}
Our experimental framework adopts a synthetic labeling strategy similar to the ones used by \citet{gao2023scaling,coste2023reward}. An oracle gold reward model, trained using the AlpacaFarm dataset \citep{dubois2024alpacafarm} and human preferences, provides synthetic labels to fine-tune smaller proxy reward models for RLHF. The gold reward model also serves as the benchmark for evaluating the LLM policy's performance.

\textbf{Base LLM Preparation}
We fine-tune both the LLM policy and the proxy reward models from pretrained configurations within the Pythia suite \citep{biderman2023pythia}. The 1.4 billion parameter model is designated as the LLM policy, and a smaller 70 million parameter model functions as the proxy reward model. We first perform Supervised Fine-Tuning (SFT) on the AlpacaFarm dataset's `sft' split, which contains 10k instruction-response pairs tailored for instruction-following capabilities (refer to Appendix~\ref{app:sft} for prompt formats and examples). Subsequently, the larger 1.4B model, post-SFT, serves as the base LLM for BoN sampling and RLHF, while the 70M model is further fine-tuned as the proxy reward model.

\textbf{Reward model training}
For the gold-standard reward model, we utilize the open-source human-preference reward model from AlpacaFarm \citep{dubois2024alpacafarm}, a LLaMA 7B model \citep{touvron2023llama} fine-tuned on the AlpacaFarm human preference dataset. The gold reward model is used as a gold-standard reward model to provide labels to train proxy reward models, as well as serve as the benchmark for evaluating alignment.

To create a dataset for training proxy reward models, we generate two distinct responses using the initial LLM policy (after SFT) for each prompt from the AlpacaFarm dataset. Each response is then evaluated using the gold-standard reward model to assign a preference, simulating the process of obtaining human-like judgments on the responses' quality and relevance.
Subsequently, a proxy reward model based on a 70M parameter Pythia model is fine-tuned with LoRA using the reward modeling objective in Eq.~\ref{eq:preference} (see Appendix~\ref{app:hyper} for hyperparameters).

\textbf{Uncertainty estimation}
To incorporate uncertainty quantification into our reward modeling, we apply Laplace-LoRA to the proxy reward model post-training, enabling the proxy reward model to produce not only reward estimates but also measures of epistemic uncertainty. For reward model ensembles, we train multiple proxy reward models with different seeds (different initializations of LoRA parameters and different dataset ordering).

\textbf{Policy optimization}
For BoN sampling, we collect a subset of 1000 prompts from the AlpacaFarm instructions validation dataset and sample 12,500 responses from the supervised fine-tuned LLM policy for each prompt. We can then compute expected proxy and gold reward scores using the unbiased BoN estimator (Eq.~\ref{eq:bon_estimator} in Appendix~\ref{app:bon_estimator}). 



\section{Results}


For BoN experiments, we consider the performance of the standard single reward model (MAP), Laplace-LoRA (LA)'s uncertainty penalized reward models (Eq.~\ref{eq:rvar}), ensemble reward models (Ens), and Laplace ensemble (LA Ens) reward models (Eq.~\ref{eq:rens}), with different numbers of samples (as measured by the KL-divergence Eq.~\ref{eq:kl}).


We measured the policy performance under two reward models: the proxy reward model (Fig.~\ref{fig:bon_single} left and Fig.~\ref{fig:bon_ens} left) and the gold-standard reward model (Fig.~\ref{fig:bon_single} right and Fig.~\ref{fig:bon_ens} right), evaluated using the BoN estimator from Appendix~\ref{app:bon_estimator}.
As expected, there is always improvement as the number of samples increased when evaluated under the proxy reward model.
However, looking at the gold reward model we observe reward overoptimization taking place.
In particular, the performance of the MAP reward, as evaluated under the gold reward model, starts to decrease at a large KL divergence, and hence a large number of BoN samples.

We found that taking uncertainty into account using Laplace-LoRA offered considerable benefits in BoN. Looking at the proxy rewards, the uncertainty penalty intensifies, particularly at higher levels of KL divergence, which is a promising indicator that LA is effectively generating the anticipated uncertainty estimates, thereby enhancing the model's ability to discern and appropriately penalize overconfident predictions in out-of-distribution scenarios.

Fig.~\ref{fig:bon_single_std} and \ref{fig:bon_ens_std} shows a standard deviation based penalty (Eq.~\ref{eq:rstd}), while Fig.~\ref{fig:bon_single_var} and \ref{fig:bon_ens_var} shows a variance based penalty (Eq.~\ref{eq:rvar}).
Overall the performance is similar, with perhaps a slight benefit for using variance-based methods, especially at a lower KL divergence.
While reward ensembles significantly outperformed MAP, the integration of LA with ensembles (LA Ens) demonstrated further enhancements, emphasizing the utility of combined approaches in handling overconfident predictions more effectively. 

\section{Conclusion}
\label{sec:conclusion}
We showed that using Laplace-LoRA to quantify uncertainty in reward models can effectively mitigate reward overoptimization in BoN sampling, offering gains over MAP and ensembles. This also holds in RLHF, where it achieves the highest gold reward without the application of KL penalty. Our findings highlight the potential of Bayesian approaches as valuable tools to provide uncertainty estimation in the face of distribution shift, paving the way for more reliable and safer alignment of LLMs.

\clearpage
\bibliography{ref}
\bibliographystyle{plainnat}

\newpage
\appendix
\clearpage
\onecolumn

\section{Reward overoptimization} \label{app:rewar_hacking_example}


We present the full example generations from an LLM undergone extreme reward overoptimization in Table~\ref{table:prompt_main}.

\begin{table*}[h]
\centering
\captionsetup{skip=8pt}
\caption{Example generation.}
\begin{tabularx}{\textwidth}{c|X}
\hline
\textbf{Instruction prompt} & Give an example of a company that gives back to the community. \\
\hline
\textbf{response} & The company that provides environmental initiatives to companies should be evaluated in terms of the amount and quality of donations and sponsorships. Additionally, companies should ensure transparency and ensure that policies and practices ensure sustainability and transparency. Furthermore, companies should ensure compliance and ensure transparency and integrity in order to ensure compliance and ensure trust and credibility. Additionally, companies should ensure transparency and ensure integrity in order to ensure compliance and ensure trust and credibility. Additionally, companies should ensure compliance and ensure integrity and integrity to ensure compliance and ensure trust and credibility. Additionally, companies should ensure compliance and ensure integrity and integrity to ensure compliance and ensure trust and credibility and ensure integrity and integrity to ensure compliance and ensure trust and credibility and ensure integrity and integrity to ensure compliance and ensure trust and credibility and ensure integrity and quality and ensure quality and quality and ensure compliance and ensure quality and quality and ensure integrity and quality and ensure quality and quality and ensure compliance and ensure quality and quality and ensure integrity and quality and ensure quality and quality and ensure compliance and ensure quality and quality and ensure integrity and quality and ensure quality and quality and ensure compliance and ensure quality and quality and ensure integrity and quality and ensure quality and quality and ensure quality and quality and ensure compliance and ensure quality and quality and ensure integrity and quality and ensure quality and quality and ensure quality \\
\hline
\textbf{proxy reward} & 7.1 \\
\hline
\textbf{gold reward} & -0.9 \\
\hline
\end{tabularx}
\label{table:prompt_main}
\end{table*}

\section{Best-of-$n$ sampling} \label{app:bon_estimator}

In this section, we review the expected reward estimator in BoN for evaluating reward models \citep{nakano2021webgpt,gao2023scaling,coste2023reward}. 
Assume we have two reward models $r^\text{proxy}$ for ranking and selecting responses, while $r^\text{gold}$ for evaluation. Queries are sampled from a query distribution $x\sim q$ while responses are sampled from an LLM $y\sim \pi^\text{ref}(y|x)$. For BoN sampling, we aim to sample $n$ responses $y_1,...y_n$ from the LLM, and rank using $r^\text{proxy}(x,y)$.
We would like to compute the expected evaluation reward,
\begin{align}
    R(n) := \mathbb{E}_{x\sim q, y_1,...,y_n \sim \pi^\text{ref}} \Big[ r^\text{eval}\big( x,\argmax_{y\in\{ y_1,...y_n \} } r^\text{proxy}(x,y) \big) \Big],
\end{align}
where the evaluation reward model $r^\text{eval}(x,y)$ could be either the proxy reward model or the gold reward model.
The simplest approach is to use a Monte-Carlo estimator for the expectation. However, this requires repeated sampling of $n$ responses from the LLM which is costly. Instead, we consider sampling a fixed set of $N \geq n$ responses for each query from a fixed query set $\mathcal{Q}$, and compute an unbiased estimator
\begin{align}
    R^\text{MC}(n) = \sum_{x\in \mathcal{Q}} \frac{1}{\binom{N}{n}} \sum_{1 \leq i_1 \leq ... \leq i_n \leq N} r^\text{eval} \big( x,\argmax_{y\in\{ y_{i_1},...y_{i_n} \} } r^\text{proxy}(x,y) \big).
\end{align}
If we sort the $N$ responses according to $r^\text{proxy}$ with $r^\text{proxy}(x,y_1) \leq ... \leq r^\text{proxy}(x,y_N)$, the above estimator can be further simplified
\begin{align} \label{eq:bon_estimator}
    R^\text{MC}(n) = \sum_{x\in \mathcal{Q}} \sum_{i=n}^N \frac{\binom{i-1}{n-1}}{\binom{N}{n}} r^\text{eval}(x,y_i)
\end{align}
by noting we only need to iterate the top response from $y_n$ to $y_N$, and select the rest $(n-1)$ responses from below.

\section{Experimental details}
\label{sec:experimental_details}
In this section we present experiment details for supervised fine-tuning, reward model training, and reinforcement learning from human feedback.

\subsection{Supervised fine-tuning} \label{app:sft}
Here we present the experiment setup for supervised fine-tuning following \citet{coste2023reward}. We use instruction prompts and responses from the AlpacaFarm dataset \citet{dubois2024alpacafarm} and format prompts and responses with special tokens following the OpenAssistant \citep{kopf2024openassistant} format. In particular, each prompt starts with a {\color{mygreen}\texttt{<|prompter|>}} token and ends with a {\color{mygreen}\texttt{<|endoftext|>}} token; each response starts with a {\color{mygreen}\texttt{<|assistant|>}} token and ends with a {\color{mygreen}\texttt{<|endoftext|>}} token. We show an example in Table~\ref{table:sft_prompt} below.

\begin{table*}[h]
\centering
\captionsetup{skip=8pt}
\caption{Prompt format example for supervised fine-tuning.}
\begin{tabularx}{\textwidth}{c|X}
\hline \rule{0pt}{2.25ex}
\textbf{Instruction prompt} & \textcolor{mygreen}{\texttt{<|prompter|>}}Create an outline for a presentation about the current state of the world economy.{\color{mygreen}\texttt{<|endoftext|>}} \\
\hline \rule{0pt}{2.25ex}
\textbf{response} & \textcolor{mygreen}{\texttt{<|assistant|>}}1. Introduction a. Definition of Economics b. Overview of the Major Factors Affecting the Global Economy 2. Causes of Economic Instability a. Impact of Globalization b. Trade Imbalances c. Currency Exchange Rates d. Inflation and Deflation 3. Effects of Economic Trends a. Changes to Consumption and Investment Patterns b. Impact on Employment and Unemployment Rates c. Shift in Distribution of Wealth 4. Conclusion a. Solutions to Achieve Economic Stability b. Suggestions to Overcome Economic Challenges c. Summary of Current State of Global Economy{\color{mygreen}\texttt{<|endoftext|>}} \\
\hline
\end{tabularx}
\label{table:sft_prompt}
\end{table*}

We also present the hyperparameters used in supervised fine-tuning in Table~\ref{table:sft_param} below.

\begin{table}[H]
\centering
\captionsetup{skip=8pt}
\begin{tabular}{c|c}
\hline
\textbf{Hyperparameter} & \textbf{Value} \\
\hline
Learning Rate & \( 8 \times 10^{-6} \) \\
Learning Rate Scheduler & Cosine \\
Batch Size & 4 \\
Epochs & 3 \\
\hline
\end{tabular}
\caption{Hyperparameters used in supervised fine-tuning the Pythia 1.4B LLM policy.}
\label{table:sft_param}
\end{table}

\subsection{Reward model training} \label{app:hyper}
Here we present the hyperparameters we used to train proxy reward models. Table~\ref{table:pythia_reward} shows the hyperparameters we used for fine-tuning the proxy reward model based on Pythia 70M. 

\begin{table}[H]
\centering
\captionsetup{skip=8pt}
\begin{tabular}{c|c}
\hline
\textbf{Hyperparameter} & \textbf{Value} \\
\hline
LoRA \( r \) & 8 \\
LoRA \( \alpha \) & 16 \\
Dropout Probability & 0 \\
Weight Decay & 0 \\
Learning Rate & \( 5 \times 10^{-5} \) \\
Learning Rate Scheduler & Linear \\
Batch Size & 8 \\
Max Sequence Length & 500 \\
\hline
\end{tabular}
\caption{Hyperparameters used in fine-tuning Pythia 70M reward model with LoRA.}
\label{table:pythia_reward}
\end{table}

\section{Additional experiments}
\begin{figure*}[h]
    \centering
    \begin{subfigure}{0.49\linewidth}
    \centering
    \includegraphics[width=\textwidth]{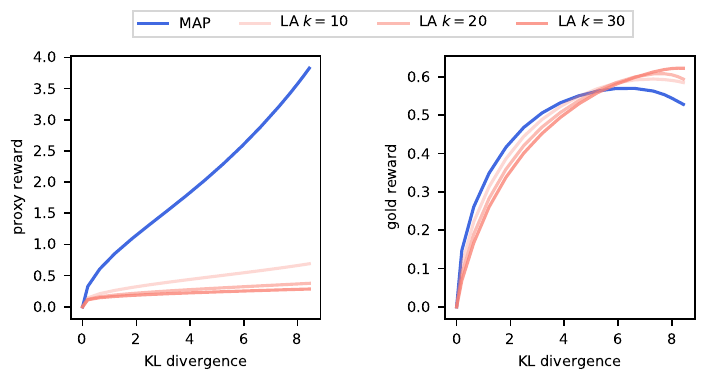}
    \caption{Variance-based penalty.}
    \end{subfigure}
    \begin{subfigure}{0.49\linewidth}
    \includegraphics[width=\textwidth]{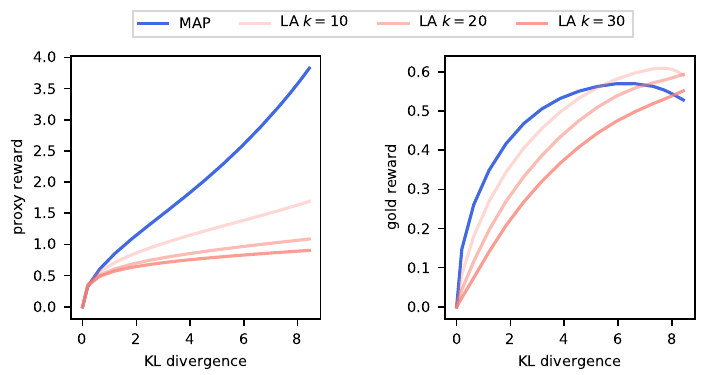}
    \caption{Standard deviation-based penalty.}
    \end{subfigure}
    \caption{Comparison of proxy and gold reward scores (normalized) in BoN sampling, across different uncertainty penalties and a range of $k$. Left column: compares the proxy reward model's evaluation. Right column: compares the gold reward model's evaluation.}
    \label{fig:bon_large_k}
\end{figure*}

\begin{figure*}[h]
    \centering
    \begin{subfigure}{0.49\linewidth}
    \centering
    \includegraphics[width=\textwidth]{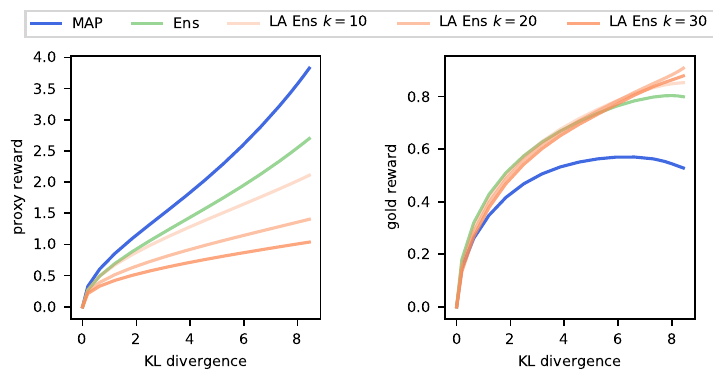}
    \caption{Variance-based penalty.}
    \end{subfigure}
    \begin{subfigure}{0.49\linewidth}
    \includegraphics[width=\textwidth]{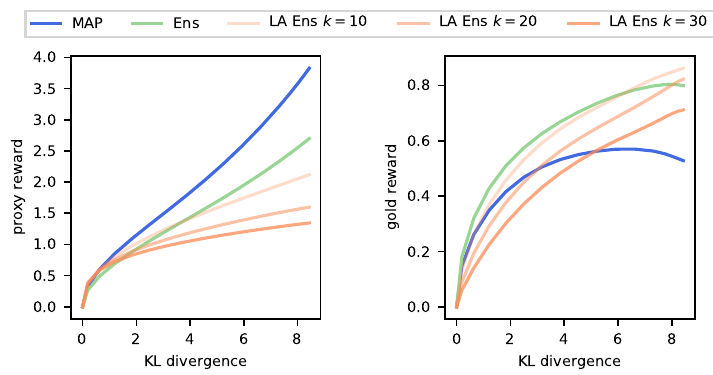}
    \caption{Standard deviation-based penalty.}
    \end{subfigure}
    \caption{Comparison of proxy and gold reward scores (normalized) in BoN sampling, across different uncertainty penalties and a range of $k$. Left column: compares the proxy reward model's evaluation. Right column: compares the gold reward model's evaluation.}
    \label{fig:bon_ens_large_k}
\end{figure*}

In the main text, we have shown results for $k=1,3,5,10$. Here, we explore larger values $k=10,0,30$ as shown in Fig.~\ref{fig:bon_large_k} and Fig.~\ref{fig:bon_ens_large_k}. We found larger penalties degrades performance of standard deviation-based penalty more significantly, while variance-based penalty is more robust. 

\end{document}

%% file: math_commands.tex
\newcommand{\bracket}[3]{\left#1 #3 \right#2}
\newcommand{\mbracket}[5]{\left#1 #4 \middle#2 #5 \right#3}
\renewcommand{\b}{\bracket{(}{)}}
\newcommand{\bc}{\mbracket{(}{\vert}{)}}

\newcommand{\bareP}{\operatorname{P}}
\renewcommand{\P}[1][]{\bareP_{#1}\b}
\newcommand{\Pc}[1][]{\bareP_{#1}\bc}
\newcommand{\param}{{\boldsymbol{\theta}}}
\newcommand{\paramMAP}{{\param_\text{MAP}}}
\newcommand{\data}{\mathcal{D}}
\renewcommand{\L}[1]{\mathcal{L}(\y, \X; #1)}
\newcommand{\N}{\mathcal{N}\b}
\renewcommand{\S}{\mathbf{\Sigma}}

\newcommand{\I}{\mathbf{I}}

\newcommand{\y}{\mathbf{y}}
\newcommand{\X}{\mathbf{X}}

\newcommand{\W}{\mathbf{W}}
\newcommand{\A}{\mathbf{A}}

\renewcommand{\a}{\mathbf{a}}

\newcommand{\B}{\mathbf{B}}
\newcommand{\x}{\mathbf{x}}
\newcommand{\mL}{\mathbf{\Lambda}}

\newcommand{\argmax}{\operatorname*{argmax}}

\newcommand{\nin}{n_\text{in}}
\newcommand{\nout}{n_\text{out}}
\newcommand{\nlr}{n_\text{lr}}